\documentclass[lettersize,journal]{IEEEtran}
\usepackage{amsmath,amsfonts}
\usepackage{algorithmic}
\usepackage{algorithm}
\usepackage{array}
\usepackage[caption=false,font=normalsize,labelfont=sf,textfont=sf]{subfig}
\usepackage{textcomp}
\usepackage{stfloats}
\usepackage{url}
\usepackage{verbatim}
\usepackage{graphicx}
\usepackage{cite}
\usepackage{booktabs}
\usepackage{hyperref}
\usepackage{multirow}
\usepackage{colortbl}
\usepackage{ulem}
\usepackage[table]{xcolor}
\hypersetup{
    colorlinks=true,
    urlcolor=magenta
}

\begin{document}

\title{MGIMM: Multi-Granularity Instruction Multimodal Model for Attribute-Guided Remote Sensing Image Detailed Description}

\author{Cong Yang, Zuchao Li, Lefei Zhang,~\IEEEmembership{Senior Member}
\thanks{Manuscript received 3 Jun 2024; revised XXX.}
\thanks{The authors are with the National Engineering Research Center for Multimedia Software, School of Computer Science, Wuhan University, Wuhan, 430072, P. R. China, and also with the Hubei Luojia Laboratory, Wuhan, 430079, P. R. China. (e-mail: \{yangcong356, zcli-charlie, zhanglefei\}@whu.edu.cn)}}

\markboth{Journal of \LaTeX\ Class Files,~Vol.~14, No.~8, August~2021}%
{Shell \MakeLowercase{\textit{et al.}}: A Sample Article Using IEEEtran.cls for IEEE Journals}


\maketitle

\begin{abstract}
  Recently, large multimodal models have built a bridge from visual to textual information, but they tend to underperform in remote sensing scenarios. This underperformance is due to the complex distribution of objects and the significant scale differences among targets in remote sensing images, leading to visual ambiguities and insufficient descriptions by these multimodal models. Moreover, the lack of multimodal fine-tuning data specific to the remote sensing field makes it challenging for the model's behavior to align with user queries. To address these issues, this paper proposes an attribute-guided \textbf{Multi-Granularity Instruction Multimodal Model (MGIMM)} for remote sensing image detailed description. MGIMM guides the multimodal model to learn the consistency between visual regions and corresponding text attributes (such as object names, colors, and shapes) through region-level instruction tuning. Then, with the multimodal model aligned on region-attribute, guided by multi-grain visual features, MGIMM fully perceives both region-level and global image information, utilizing large language models for comprehensive descriptions of remote sensing images. Due to the lack of a standard benchmark for generating detailed descriptions of remote sensing images, we construct a dataset featuring 38,320 region-attribute pairs and 23,463 image-detailed description pairs. Compared with various advanced methods on this dataset, the results demonstrate the effectiveness of MGIMM's region-attribute guided learning approach. Code can be available at \href{https://github.com/yangcong356/MGIMM.git}{https://github.com/yangcong356/MGIMM.git}
\end{abstract}


\section{Introduction}
\IEEEPARstart{I}{n} the field of remote sensing \cite{ZZD2016, XZD2022, ZZ2022}, describing remote sensing images with textual information is a task that translates discrete image information into text, enabling decision-makers to understand the content of images more intuitively \cite{SZ2017}. Therefore, remote sensing image captioning (RSIC) is crucial for urban planning, military intelligence decision-making, and more \cite{WHZL2021, LW2024, YL2024}. Despite the significant progress made by existing vision-language joint models in the domain of RSIC, these methods can only generate single and simple sentence descriptions. This inevitably leads to the models overlooking some of the geographical information contained within remote sensing images. A prime illustration of this is how descriptions of remote sensing images tend to offer only a broad outline of the key geographical features within the images. They lack detailed insights into the exact positions of these critical features \cite{YYH2021}, the dynamics between various key elements, and how these features interact with their surroundings. Therefore, utilizing coherent and accurate detailed descriptions to convey both global and local visual information in remote sensing images is increasingly important. However, generating a description with detailed geographical information is not an easy task, even for humans.

An information-rich paragraph should start with key sentences that capture the essence of the remote sensing image scene, followed by supportive sentences that include concepts of actions, attributes, and relationships among visual objects\cite{XZDY2022, CFX2020, NF2022}. Recently, driven by advancements in generative artificial intelligence, large multimodal models like LLaVA \cite{LLL2023}, InstructBLIP \cite{DLL2023}, LLaMA-Adapter v2 \cite{GHZ2023}, and MiniGPT-v2 \cite{CZS2023} have significantly advanced the capabilities of vision-language generation tasks. These multimodal models utilize large language models as a universal interface to construct multimodal frameworks with robust image-text understanding capabilities. In these large multimodal models, the feature space for specific tasks is tuned to align with the feature space of pre-trained large language models, allowing for the output of user-specified generated content. Thus, leveraging artificially set visual instruction tuning to guide large language models in providing detailed descriptions of remote sensing images has become possible. However, existing large multimodal models generally lack domain-specific knowledge \cite{LMX2022, DYL2022, ZJB2023} of the remote sensing field, meaning the complex background information of remote sensing images, significant scale variations of geographical targets, and the absence of precise remote sensing image-text descriptions hinder these models' ability to excel in vision-language understanding within the remote sensing domain. Moreover, existing large multimodal models inherently lack regional awareness capabilities \cite{PWD2023, PGD2023, ZSC2023}, which may lead to visual ambiguities when these models perceive remote sensing images with complex geographical object distributions. Specifically, this could result in geographical targets with significant scale variations being aligned with incorrect attribute descriptions.

To enable multimodal models to eliminate visual ambiguities and gain a preliminary understanding of the remote sensing domain, we propose the Multi-Granularity Instructional Multimodal Model (MGIMM). The core design philosophy of this model is to equip the multimodal models with region-level attribute alignment capabilities through a two-stage attribute-guided learning paradigm, thereby enabling them to provide detailed global descriptions of remote sensing images. Specifically, MGIMM first uses region-level instruction and region-level bounding boxes to force the multimodal model to align geographical targets with their corresponding attribute descriptions. Then, MGIMM employs designed visual instructions to allow the large language model to fully perceive the remote sensing images within the set rules of the instructions and complete detailed descriptions of the images. Furthermore, due to the lack of attribute-guided detailed description datasets in the remote sensing domain, this article builds on the DIOR-RSVG dataset, which has 38,320 region-attribute pairs. Experts in the remote sensing field annotate each image in the DIOR dataset \cite{CWL2022} with 23,463 high-quality image-detailed description pairs. This dataset is referred to as the remote sensing image detailed description (DIOR-IDD) Dataset, and it is used to train MGIMM.

In summary, our work has three main contributions:
\begin{itemize}
    \item This paper presents MGIMM, a progressive learning model that evolves from region-level to image-level instruction tuning. By aligning region-attribute descriptions of geographic targets, region-level instruction tuning addresses visual ambiguities from large-scale variations in remote sensing images. Image-level instruction tuning then enables the large language model to grasp the remote sensing domain, leveraging its capacity for a long-text generation to deliver detailed remote sensing image descriptions.
    \item During the region instruction tuning phase, to acquire regional image features through bounding boxes, we design a region interactive module that facilitates interaction between bounding boxes and global image features. This region interaction module is adept at aligning geographic target area-attribute descriptions.
    \item To facilitate model training and evaluation, we construct DIOR-IDD, a detailed description dataset for remote sensing images that includes 23,463 image-text pairs. Additionally, to achieve attribute-guided detailed descriptions of remote sensing images, we combine DIOR-RSVG and DIOR-IDD to form the training and evaluation dataset required by MGIMM.
\end{itemize}

The structure of this paper is as follows: in the "Related Work" section, we discuss relevant works, including multimodal understanding in remote sensing and multimodal instruction tuning. In the "Dataset Construction" section, we detail the process and principles used for constructing the dataset. In the "Methodology" section, we present a detailed introduction to the proposed method. In the "Experiments" section, we demonstrate the effectiveness of our approach through comprehensive comparative experiments, ablation studies, and convergence analysis. Finally, in the "Conclusion" section, we provide a summary and outlook on the problem.

\section{Related Works}
\subsection{Remote Sensing Multimodal Understanding}
Recently, remote sensing image-text multimodal understanding tasks have garnered significant attention \cite{PMB2023, SFL2022, ZSZ2022, KM2023, ZSZ2024, SBC2024}. The focus of image-level multimodal understanding tasks is primarily on RSIC. The pioneering work in this direction was conducted by Shi et al. \cite{SZ2017}, who first highlighted that the main challenge in RSIC lies in the model's need to perceive semantic information of various geographical targets across different scales and express it in complete sentences. Subsequently, considerable research \cite{LZGL2022, WHZL2022, ZZYG2022} has been dedicated to addressing the multiscale problem of remote sensing geographic targets, propelling the development of image-level multimodal understanding tasks. Inspired by large language models' powerful text generation capabilities, existing remote sensing image works \cite{MRD2022, GSF2022, LZC2022, LZC2023} have begun to leverage these models to align image-level representations with corresponding text descriptions. They all align image-text information through a visual-textual interaction layer, connecting the visual encoder and the large language model to facilitate the generation of text from remote sensing images. Additionally, remote sensing visual grounding (RSVG) is an emerging region-level multimodal understanding task that involves providing bounding boxes for specific objects of interest using remote sensing images and related query statements. Sun et al. \cite{SFL2022} have proposed a corresponding dataset for the RSVG task and designed an adaptive regional attention fusion module that integrates geographic target attribute information with remote sensing image information, following the typical design of remote sensing multimodal understanding tasks. Another similar work \cite{ZXY2023} filters out irrelevant noise information in cluttered backgrounds of remote sensing images, achieving precise representations of queried image regions with geographic target attribute descriptions. However, neither image-level nor region-level multimodal understanding tasks have been able to simultaneously eliminate visual ambiguities and provide detailed descriptions of both local geographical target information and global image environmental information.

\subsection{Multimodal Instruction Tuning}
In the field of natural language processing \cite{HXJ2023, RLB2023}, researchers have explored methods for instructing large language models like GPT-3 \cite{chatgpt} and OPT \cite{ZRG2022} to follow various command descriptions to solve any problem, leading to the development of mature instruction-tuned large language models such as InstructGPT \cite{OWA2022}, ChatGPT\cite{chatgpt}, and LLaMA \cite{TLI2023}. The results show that this simple adjustment strategy enables large language models to effectively follow predefined instructions and support new, unseen instructions, significantly improving performance in zero-shot scenarios and greatly enhancing the models' generalization abilities \cite{WGC2023, LHV2023, SWZ2023}. Consequently, LLaVA and Instruct-BLIP naturally introduced this concept into the multimodal information processing field, fine-tuning multimodal models with synthesized multimodal instruction adjustment data to enhance their ability to follow instructions. Thanks to the significant performance improvement from the multimodal instruction adjustment strategy, existing work \cite{ADL2022, LLS2023} either directly aligns external modal embeddings with large language models or uses expert models to translate external modalities into natural language that large language models can process. Then, it injects multimodal information into large language models, treating them as powerful reasoners. However, these methods mainly focus on image-level coarse-grained tasks and are ineffective for fine-grained multimodal perception problems. Recently, inspired by the segment anything model \cite{KMR2023}, Vision-LLM \cite{WCC2023} and MiniGPT-v2 \cite{CZS2023} further utilized instruction tuning data to address fine-grained visual perception tasks such as visual localization and object detection. These methods demonstrate the potential of fully exploiting large language models' understanding of multimodal instructions through instruction-tuning strategies. Based on this idea, our approach aims to use a dual-guided process of region-level instruction tuning and image-level instruction tuning in a dual-guided process to enable multimodal models to fully understand and perceive remote sensing images, thereby eliminating visual ambiguities caused by the scale variation of geographical targets.

\section{Dataset Construction}
To eliminate visual ambiguities in detailed descriptions of remote sensing images, we constructed an attribute-guided remote sensing image description dataset (Attribute-Guided DIOR-IDD) based on the DIOR object detection dataset. In this section, we provide a detailed overview of the construction process of the attribute-guided DIOR-IDD dataset.

\subsection{DIOR-RSVG}
Zhan et al. \cite{ZXY2023} construct a large-scale DIOR-RSVG dataset consisting of 17,402 remote sensing images and 38,320 unique attribute descriptions. The DIOR-RSVG dataset is constructed based on the DIOR object detection dataset, with incorrect bounding boxes removed. In this dataset, each object instance in a remote sensing image corresponds to a distinct language expression. The dataset includes 20 object categories, with the average length of expressions being 7.47 words and a vocabulary size of 100. Vehicles are the most common category, while harbors are the least common. The remaining 18 categories each account for a relatively uniform proportion, ranging from 2\% to 10\%. Furthermore, the most common object attributes include color, size, and absolute position. Additionally, the most common relationship attributes pertain to relative location, such as upper right and left. The seven main attributes of DIOR-RSVG are shown in Table \ref{tab1}.

\begin{table}[!h]
\centering
\caption{Specific Information for Each Attribute.}
\begin{tabular}{ll}
\hline
\textbf{Attribute} & \textbf{Example} \\
\hline
category & (e.g. “vehicle, harbor”) \\
color & (e.g. “green, blue”) \\
size & (e.g. “tiny, big”) \\
geometry & (e.g. “square, round”) \\
absolute location & (e.g. “top-left of the image”) \\
relative location relation & (e.g. “the plane is left of the terminal”) \\
relative size relation & (e.g. “the vehicle is smaller than the tree”) \\
\hline
\end{tabular}
\label{tab1}
\end{table}

\begin{figure}[!ht]
    \centering
    \begin{tabular}{p{3.5cm} p{3.5cm}}
         \includegraphics[width=3.5cm]{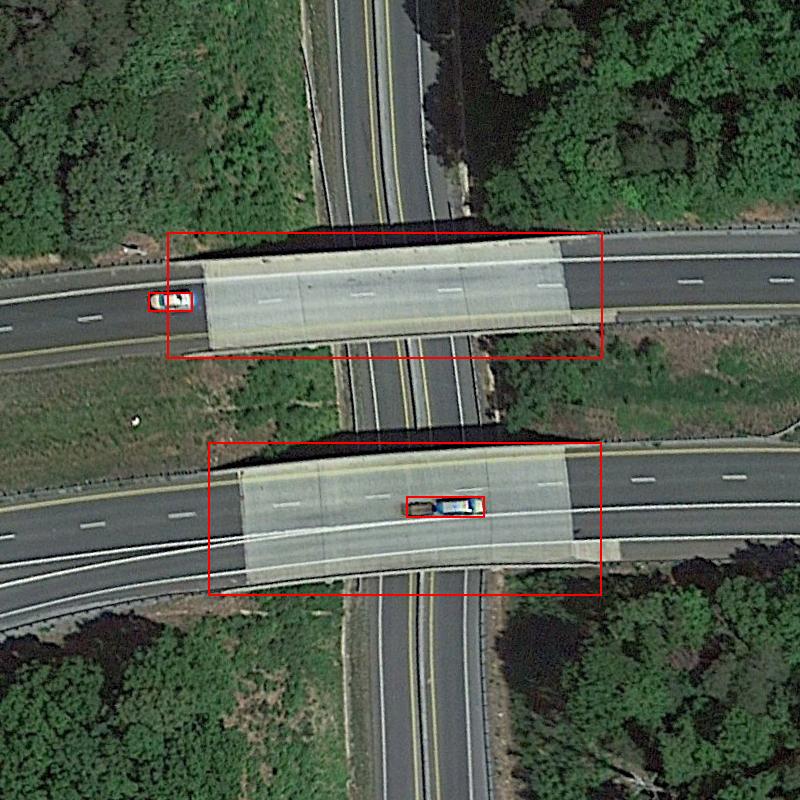}& \includegraphics[width=3.5cm]{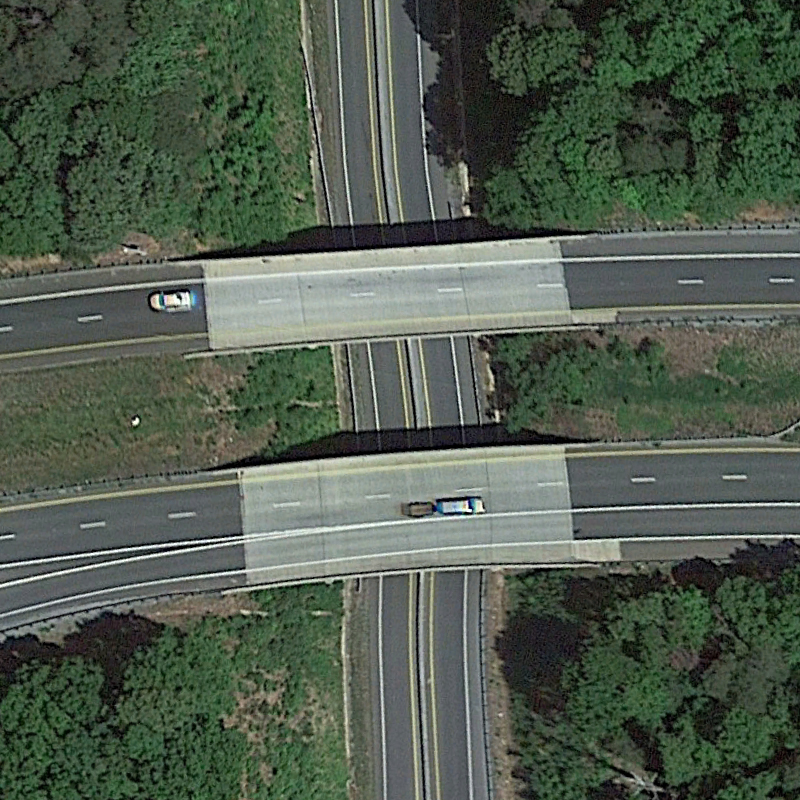} \\
         \textbf{Region-level Instruction}: Based on the provided region of the remote sensing image, describe the basic attributes of the main objects in that region. &         \textbf{Image-level Instruction}: Illustrate the remote sensing image through a descriptive explanation. \\
        \textbf{bbox}: [208.0, 442.0, 393.0, 153.0] \newline
        \textbf{Attribute}: The overpass is on the left of the small vehicle \newline
        \textbf{bbox}: [167.0, 232.0, 435.0, 126.0] \newline
        \textbf{Attribute}: The overpass is on the right of the vehicle on the left \newline
        \textbf{bbox}: [148.0, 292.0, 44.0, 19.0] \newline
        \textbf{Attribute}: A tiny vehicle \newline
        \textbf{bbox}: [406.0, 496.0, 78.0, 21.0] \newline
        \textbf{Attribute}: A small vehicle &
        \textbf{Caption}: From a satellite perspective, this image shows an overpass and vehicles, with the position of the overpass clearly visible. The overpass is situated in the center of the image, with two ends connected to a road that runs through the left and right sides. Underneath the overpass, there is only one road visible, spanning from the top to the bottom of the picture. On the road above the overpass, several vehicles can be seen driving. \\
    \end{tabular}
    \caption{Example Display of the Attribute-Guided DIOR-IDD Dataset.}
    \label{fig1}
\end{figure}

\subsection{DIOR-IDD}
\subsubsection{Basic Data}
Since the DIOR-IDD dataset requires a large number of remote sensing images with different object bounding boxes and their corresponding attribute descriptions, we choose the DIOR \cite{CWL2022} object detection dataset as the benchmark for annotated detailed descriptions. Additionally, we combine the annotated detailed description dataset with the DIOR-RSVG dataset, which contains numerous images with different object bounding boxes and their corresponding attribute descriptions, to create the attribute-guided DIOR-IDD dataset proposed in this paper.

The DIOR dataset includes 23,463 remote sensing images, each sized at 800 × 800 pixels, with a spatial resolution range from 0.5m to 30m. First, two remote sensing experts professionally annotated the remote sensing images. The detailed description procedure for the remote sensing images follows these principles: (1) The description should be relevant to the image content and should not contain errors or unrelated information. (2) The description should cover the main objects, attributes (such as color, shape, and texture), actions, and scenes in the image. (3) The description should use appropriate and accurate vocabulary and grammar, with proper punctuation and capitalization, while employing coherent and logical sentence structures to avoid spelling mistakes, grammatical errors, repetition, or redundancy. Based on these principles, we generate 23,463 high-quality remote sensing image-description text pairs.

\subsection{Attribute-Guided DIOR-IDD}
The attribute-guided DIOR-IDD dataset is a combination of the DIOR-RSVG and DIOR-IDD datasets. Since DIOR-RSVG is used exclusively for region-level instruction tuning, we select images from DIOR-RSVG that align with the training and validation indices of DIOR-IDD for our training data. Figure \ref{fig1} illustrates an example from the attribute-guided DIOR-IDD dataset. The text description provides a rich and detailed depiction of the scene, offering valuable references for users interacting with remote sensing knowledge. Additionally, the ‘bbox’ and ‘attribute’ on the left side of Figure \ref{fig1} correspond one-to-one.

\begin{figure*}[h]
    \centering
    \includegraphics[width=0.72\linewidth]{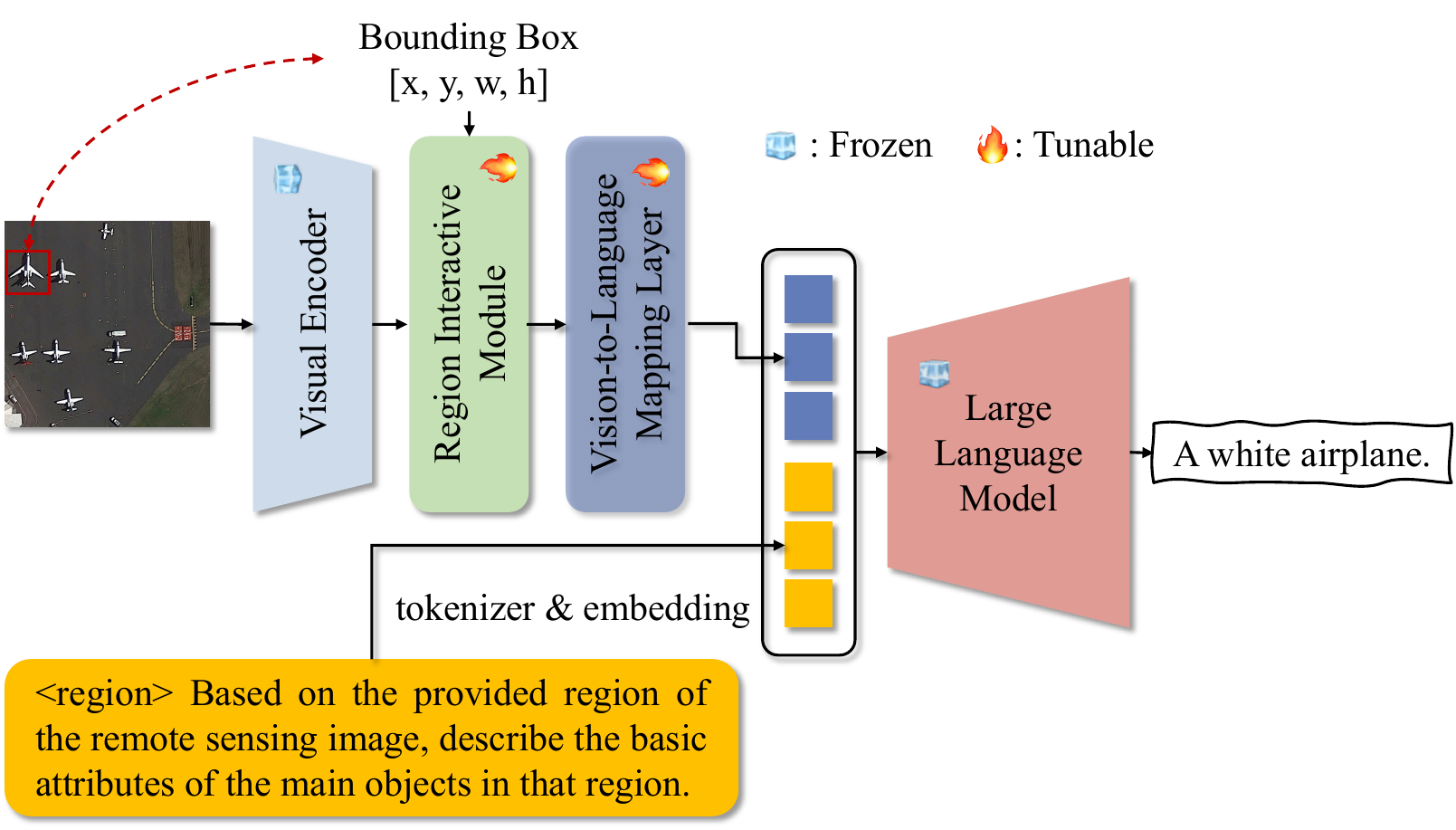}
    \caption{Region-level instruction tuning framework. Remote sensing images first pass through a visual encoder to obtain global image features. These features and bounding boxes then go through a region interactive module to obtain regional interactive features. The regional interactive features are mapped into the text space through the vision-to-language mapping layer and then concatenated with the encoded region-level instruction embeddings. The concatenated features are input into a frozen large language model to output the attribute descriptions corresponding to the geographic targets.}
    \label{fig2}
\end{figure*}

\begin{figure}[!ht]
    \centering
    \includegraphics[width=0.7\linewidth]{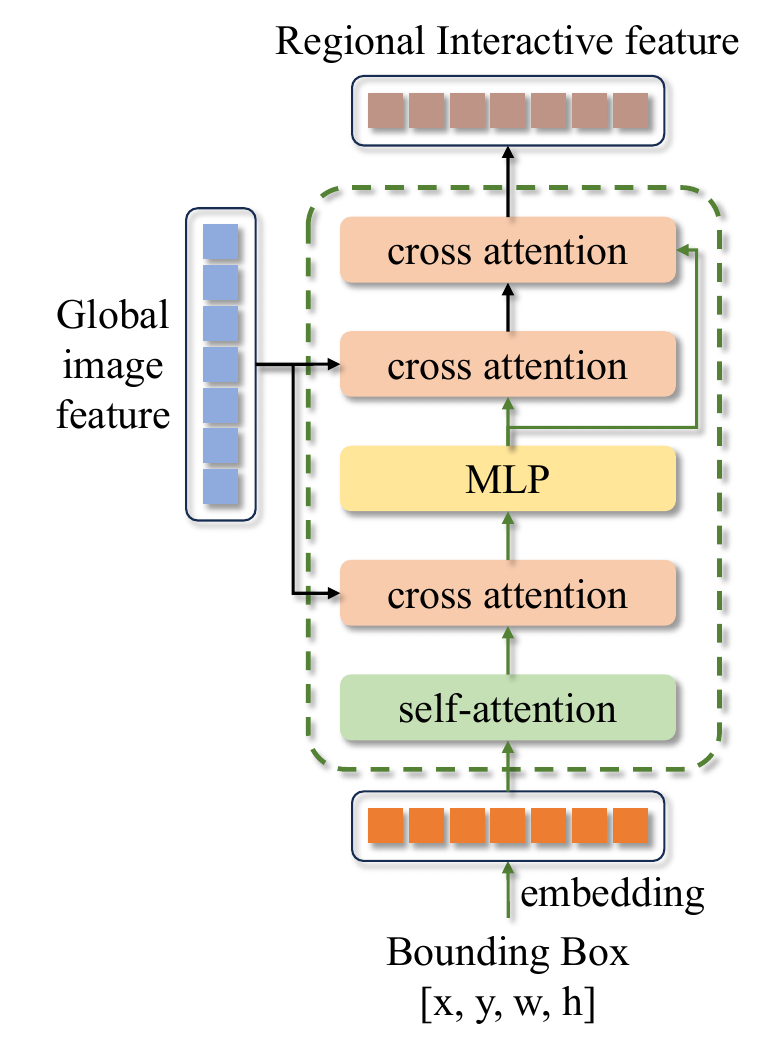}
    \caption{Overall structure of the region interactive module.}
    \label{fig3}
\end{figure}

\section{Method}
\label{sec3}
In this section, we introduce the architecture of the Multi-Granularity Instruction Multimodal Model designed for the attribute-guided detailed description of remote sensing images. This approach leverages the powerful representation and reasoning capabilities of visual models and large language models through both region-level and image-level instruction tuning. It achieves guided learning from geographic target region-attribute alignment to a comprehensive understanding of the entire remote sensing image. To implement region-level instruction tuning, we design a region interaction module that directs the model’s focus solely on the geographic information within the designated bounding boxes, effectively addressing visual ambiguities caused by significant scale variations. Additionally, we develop distinct instructions for both region-level and image-level instruction tuning to better guide the model in concentrating on specific fine-tuning tasks at different stages.

\subsection{MGIMM Architecture}
MGIMM is a two-stage multimodal understanding framework that includes a region-level instruction tuning stage for geographic target area-attribute alignment and an image-level instruction tuning stage for global image perception description. Both the region-level instruction tuning stage and the image-level instruction tuning stage encompass a visual encoder $F_{I}$, a vision-to-language mapping layer $F_{v2l}$, and a large language model $F_{llm}$. Unlike the image-level instruction tuning stage, the region-level instruction tuning stage additionally employs a region interactive module $F_{rim}$. This module queries the most relevant area features within the image through bounding boxes, compelling the multimodal model to forcibly align geographic target area features with their attribute descriptions. These components collectively facilitate MGIMM's region-guided learning paradigm, which will be detailed for all modules below.

\textbf{Visual Encoder $F_{I}$ :} Considering the outstanding contrastive language-image pre-training (CLIP) strategy and the powerful generalization ability, we adopt the CLIP-ViT-L/14@336 model developed by OpenAI as MGIMM's visual. We utilize the grid feature map from the penultimate Transformer layer of this model to represent the entire remote sensing image. This visual encoder is capable of processing input remote sensing images $\mathbf{x}_{img}$ at a relatively high resolution $336 \times 336$, obtaining 1024 dimensional high-quality global remote sensing image features $\mathbf{x}_{g}$:

\begin{equation}\label{eq1}
    \mathbf{x}_{g} = F_{I}\left(\mathbf{x}_{img}\right).
\end{equation}

\textbf{Region Interactive Module (RIM) $F_{rim}$ :} To facilitate interaction between input bounding boxes and global image features $\mathbf{x}_{g}$ for acquiring regional features $\mathbf{x}_{r}$, we utilize a dual embedding encoding approach for representing bounding boxes. This involves creating a pair of embeddings for each bounding box: the first combines a position encoding for the top-left corner with a learned embedding specifically representing the "top-left corner." The second embedding follows a similar structure, but it incorporates a learned embedding for the "bottom-right corner." This method ensures a comprehensive representation that captures both the specific locations and the conceptual roles of bounding box corners in relation to the global image context. To ensure the bounding box embeddings can interact with global image features, the bounding boxes are mapped into a 1024-dimensional learnable vector embedding $\mathbf{t}_{bbox}$.

As illustrated in Figure \ref{fig3}, the lightweight region interactive module effectively utilizes bounding box embeddings to query the most relevant regional features within the global image features. In each region interaction layer, the process starts with calculating self-attention on the embeddings of bounding boxes. This is followed by a cross-attention step where the bounding box embeddings serve as query vectors to interact with the global image embeddings. Next, every token undergoes updates through a point-wise MLP. The process then moves to another cross-attention step, but this time with the roles reversed: image embeddings act as query vectors against the bounding box embeddings. The final step involves updating the global image features using information from the bounding box embeddings. During the region-level instruction adjustment stage, we utilize the learned bounding box embedding vectors for subsequent attribute information generation. The calculation process for self-attention (SA) and cross-attention (CA) in this procedure is as follows:

\begin{small}
    \begin{equation}
        \begin{aligned}
            \mathbf{Q}_{sa} &= \mathbf{t}_{bbox} \mathbf{W}_{sa}^Q,     \mathbf{K}_{sa} = \mathbf{t}_{bbox} \mathbf{W}_{sa}^K, \mathbf{V}_{sa} = \mathbf{t}_{bbox} \mathbf{W}_{sa}^V, \\
            \mathbf{t}_{sa} &=\text{SA}(\mathbf{Q}_{sa}, \mathbf{K}_{sa}, \mathbf{V}_{sa}) = \text{softmax}\left(\frac{\mathbf{Q}_{sa} \mathbf{K}_{sa}^T}{\sqrt{{d_k}_{sa}}}\right)\mathbf{V}_{sa},
        \end{aligned}
    \end{equation}
\end{small}

\begin{small}
    \begin{equation}
        \begin{aligned}
            \mathbf{Q}_{ca} = \mathbf{x}_{g} \mathbf{W}_{ca}^Q,     \mathbf{K}_{ca} = \mathbf{t}_{sa} \mathbf{W}_{ca}^K, \mathbf{V}_{ca} = \mathbf{t}_{sa} \mathbf{W}_{ca}^V, \\
            \text{CA}(\mathbf{Q}_{ca}, \mathbf{K}_{ca}, \mathbf{V}_{ca}) = \text{softmax}\left(\frac{\mathbf{Q}_{ca} \mathbf{K}_{ca}^T}{\sqrt{{d_k}_{ca}}}\right)\mathbf{V}_{ca},
        \end{aligned}
    \end{equation}
\end{small}

\noindent where weight matrices $\mathbf{W}_{sa}^Q$, $\mathbf{W}_{sa}^K$, and $\mathbf{W}_{sa}^V$ are learnable parameters in the self-attention calculation process. Similarly, in the cross-attention calculation process, $\mathbf{W}_{ca}^Q$, $\mathbf{W}_{ca}^K$, and $\mathbf{W}_{ca}^V$ are are also learnable weight matrices. $\sqrt{{d_k}_{sa}}$ and $\sqrt{{d_k}_{ca}}$ are the dimensions of $\mathbf{K_{sa}}$ and $\mathbf{K_{ca}}$, respectively. These values are used to scale the dot product and prevent it from becoming too large in high dimensions, which can lead to the gradient vanishing problem. The softmax function is applied to each row to normalize the weights into a valid probability distribution. Overall, the process of obtaining regional interactive features $\mathbf{x}_{r}$ can be expressed as follows:

\begin{equation}
    \mathbf{x}_{r} = F_{rim}\left(\mathbf{t}_{bbox}, \mathbf{x}_{g}\right)
\end{equation}

To ensure the region interaction module can access crucial geometric information, position encoding is added to the global image features when participating in the attention layer. Furthermore, whenever the original bounding box embeddings participate in the attention layer, the original bounding box embeddings (including their position encodings) are re-added to the updated tokens. The embedding dimension in the region interactive module matches the bounding box embeddings at 1024. The output dimension of the MLP layer in the region interactive module is 2048, and the MLP layer is only applied to bounding box embeddings. In the self-attention and cross-attention layers, we set the channel dimension of queries, keys, and values to 128 to accelerate computational efficiency. All attention layers utilize 8 heads.

\textbf{Vision-to-Language Mapping Layer $F_{v2l}$ :} To efficiently map image features into the language domain, we use two simple linear projection layers for this operation. The vision-to-language mapping layer processes regional image features during the region-level instruction tuning stage. In contrast, the image-level instruction tuning stage is where the vision-to-language mapping layer handles global image features. These two processes are represented as follows:

\begin{equation}\label{eq2}
    \begin{aligned}
        \mathbf{x}_{r2l} &= F_{v2l}\left(\mathbf{x}_{r}\right), \\
        \mathbf{x}_{g2l} &= F_{v2l}\left(\mathbf{x}_{g}\right),    
    \end{aligned}
\end{equation}

\noindent where $\mathbf{x}_{r2l}$ represents the features of regional image features $\mathbf{x}_{r}$ after being mapped to the language domain, while $\mathbf{x}_{g2l}$ denotes the features of global image features $\mathbf{x}_{g}$ following their mapping to the language domain.

\textbf{Large Language Model $F_{llm}$ :} To describe geographic target regional attributes at the region-level instruction tuning stage, and to provide detailed descriptions of the entire remote sensing image at the image-level instruction tuning stage, we've opted for open-source large language models (LLaMA-2 and Phi-2) as the unified interface for various visual instruction tuning inputs. The large language models then generate corresponding descriptions of geographic object regional attributes and detailed remote sensing image descriptions at the region-level instruction tuning stage and image-level instruction tuning stage, respectively:

\begin{equation}\label{eq3}
    \begin{aligned}
           y_{rit} &= F_{llm}\left(\text{Concat}\left[\mathbf{x}_{r}, \mathbf{y}_{ri}\right]\right), \\
    y_{iit} &= F_{llm}\left(\text{Concat}\left[\mathbf{x}_{g},\mathbf{y}_{gi}\right]\right), 
    \end{aligned}
\end{equation}

\noindent where $\mathbf{y}_{ri}$ and $y_{rit}$ represent the regional instruction embeddings and description of regional geographic target attributes during the region-level instruction tuning stage, respectively. $\mathbf{y}_{gi}$ and $y_{iit}$ also signify the image-level instruction embeddings and detailed descriptions of the entire remote sensing image during the image-level instruction tuning stage, respectively.

\begin{figure*}[!ht]
    \centering
    \includegraphics[width=0.85\linewidth]{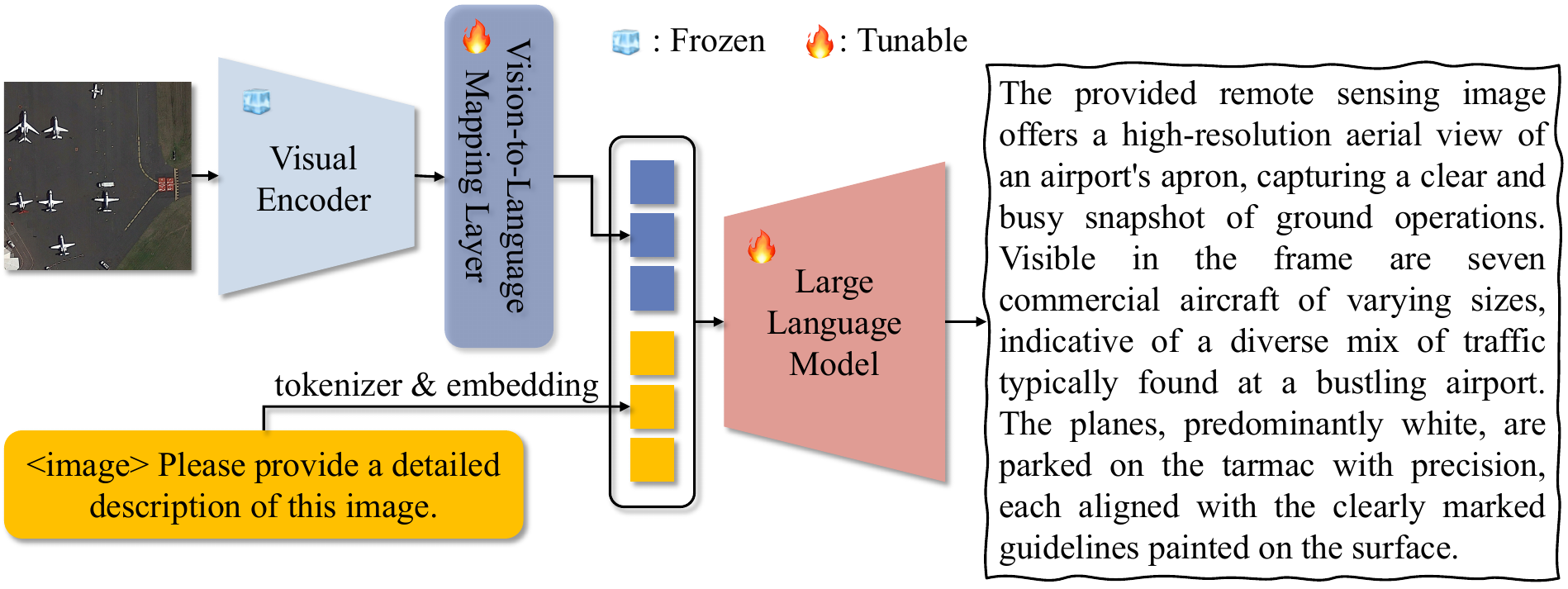}
    \caption{Image-level instruction tuning framework. First, the remote sensing image passes through a visual encoder and a vision-to-language mapping layer to obtain global image features. Then, the encoded image-level instructions are concatenated with the global image features and input into a trainable large language model to obtain a detailed description of the remote sensing image.}
    \label{fig4}
\end{figure*}

\subsection{Region-Level Instruction Tuning}
The model learns from a single, paired region-attribute description in the region-level instruction tuning process illustrated in Figure \ref{fig2}. The main goal of the region-level instruction tuning stage is to guide the model in aligning the area of the geographic target with its corresponding attribute description, thereby resolving visual ambiguity issues caused by significant scale variations in remote sensing images. At this stage, we initially load the pre-trained weights of the visual encoder and the large language model from the HuggingFace website. We only keep the region interactive module and the vision-to-language mapping layer trainable to alleviate visual ambiguity between geographic targets and their attribute texts. To enable MGIMM to focus on and respond to regional features effectively, we've expanded the vocabulary with the specialized token "\textit{\textless region\textgreater}." Based on this specific "\textit{\textless region\textgreater}" token, we've designed an instruction for the region-level instruction tuning stage, such as "\textit{\textless region\textgreater Based on the provided region of the remote sensing image, describe the basic attributes of the main objects in that region}." Before the large language model makes corresponding predictions, the "\textit{\textless region\textgreater}" specialized token in the regional instruction is replaced with image region features $\mathbf{x}_{r}$ extracted by the region interactive module. Given the regional instruction, the optimization objective function for the region-level instruction tuning stage is as follows:

\begin{equation}\label{eq4}
    \begin{aligned}
    \mathcal{L} & =\log P\left(y_{ra} \mid \text{Concat}\left[\mathbf{x}_{r}, \mathbf{y}_{ri}\right] ; \theta\right) \\
    & =\sum_{l=1}^L \log P\left(y_l \mid \text{Concat}\left[\mathbf{x}_{r}, \mathbf{y}_{ri}\right], y_{ra,<l} ; \theta\right),
    \end{aligned}
\end{equation}

\noindent where $y_{ra}$ represents the regional attribute description with a length of L. $P$ and $\theta$ stand for the conditional probability and the parameters of the large language model, respectively. $y_{ra,<l}$ is the answer token preceding the currently predicted token $y_l$.

\subsection{Image-Level Instruction Tuning}
After region-level instruction tuning, MGIMM has gained the ability to perceive regions. The image-level instruction tuning (as shown in Figure \ref{fig4}) builds on this foundation to further guide the model in fully understanding the spatial distribution of geographic targets within the remote sensing images. It leverages large language models to generate detailed descriptions of remote sensing images. In the second stage, we freeze the pre-trained visual encoder weights and train the weights of the vision-to-language mapping layer and the large language model. Both the visual encoder and the large language model weights are loaded from the HuggingFace website. The weights for the vision-to-language mapping layer are loaded from those trained during the region-level instruction tuning stage. Since the weights of the vision-to-language mapping layer trained during the region-level instruction tuning stage have already been aligned with region-attribute pairs, we do not use the region interactive module in this phase. Accordingly, to facilitate MGIMM's perception of the entire remote sensing image, we expand the vocabulary with the specialized token "\textit{\textless image\textgreater}" and have designed 15 types of image-level instructions for detailed descriptions of remote sensing images, as shown in Table \ref{tab2}. We randomly select an image-level instruction for each detailed description $y_{ga}$ of a remote sensing image with length $M$. The optimization objective function for the image-level instruction tuning is as follows:

\begin{equation}\label{eq5}
    \begin{aligned}
    \mathcal{L} & =\log P\left(y_{ga} \mid \text{Concat}\left[\mathbf{x}_{g},\mathbf{y}_{gi}\right] ; \theta\right) \\
    & =\sum_{m=1}^M \log P\left(y_m \mid \text{Concat}\left[\mathbf{x}_{g},\mathbf{y}_{gi}\right], y_{ga,<m} ; \theta\right),
    \end{aligned}
\end{equation}

\noindent where $P$ and $\theta$ stand for the conditional probability and the parameters of the large language model, respectively. $y_{ga,<m}$ is the answer token preceding the currently predicted token $y_m$.

\begin{table}[ht]
\centering
\caption{The 15 types of instructions designed for image-level instruction tuning}
\label{tab2}
\begin{tabular}{p{8cm}}
\toprule
Describe the following remote sensing image in detail.                        \\ 
Provide a detailed description of the given remote sensing image.             \\
Elaborate on the remote sensing image you see.                                \\
Share a comprehensive overview of the presented remote sensing image.         \\
Conduct a thorough analysis of the remote sensing image.                      \\
Explain the various aspects of the remote sensing image before you.           \\
Clarify the contents of the displayed remote sensing image with great detail. \\
Characterize the remote sensing image using a detailed description.           \\
Break down the elements of the remote sensing image in detail.                \\
Walk through the important details of the remote sensing image.               \\
Portray the remote sensing image with a rich, descriptive narrative.          \\
Narrate the contents of the remote sensing image with precision.              \\
Analyze the remote sensing image in a comprehensive and detailed manner.      \\
Illustrate the remote sensing image through a descriptive explanation.        \\
Write an exhaustive depiction of the given remote sensing image.             \\
\bottomrule
\end{tabular}
\end{table}

\section{Experiments}
\subsection{Implementation Details}
In Section \ref{sec3}, we provide a detailed overview of the parameter settings for the visual encoder, region interactive module, vision-to-language mapping layer, and language decoder used by MGIMM, as well as their weight-loading mechanisms. In subsequent experiments, Phi-2 \footnote{\href{https://huggingface.co/microsoft/phi-2}{Pre-trained weights for Phi-2 2.3B}} with 2.3 billion parameters proposed by Microsoft, and Llama-2 with 7 billion parameters proposed by MetaAI will both serve as the language decoders for MGIMM. We employ the AdamW optimizer with a cosine learning strategy to train the model. In the initial tuning phase, we configure the global batch size at $4 \times 64$, utilizing four RTX 4090 GPUs for training over 10 epochs. The optimizer's global learning rate during this phase is set to 1e-4. Progressing to the second tuning phase, we adjust the global batch size to $4 \times 4$ and train for 5 epochs. We further refine our approach by setting the optimizer's global learning rate to 2e-5, with a specific emphasis on the vision-to-language mapping layer, for which we assign a learning rate of 2e-6. Additionally, we use a strategy based on Low-Rank Adaptation (LoRA) \cite{HSW2022}, approximating the original large language model weight matrices in each linear layer with two low-rank weight matrices. LoRA ensures faster training speeds and prevents the loss of original knowledge embedded in the large language models trained and fine-tuned according to general natural language instructions. We set the ranks of the two low-rank matrices to 128 and 256, respectively.

\subsection{Experimental Datasets}
In the initial fine-tuning stage, we trained the MGIMM model using data from DIOR-RSVG that matches the image indices of the DIOR-IDD training set, where each object instance in the remote sensing images is associated with a unique attribute description. In the second adjustment stage, we trained the MGIMM model with the detailed remote sensing image description dataset, DIOR-IDD \footnote{\href{https://1drv.ms/f/s!AvupDOrrrLbsg8oRd20DNgwL9NWKTQ}{Dataset download link}}, which consists of 23,463 image-description pairs. These detailed descriptions have an average length of 445.50 words. Additionally, we combined the training and validation sets of the original DIOR \cite{CWL2022} dataset to form the training set of DIOR-IDD, ensuring the same proportion for the test set division. As a result, the DIOR-IDD dataset has 11,725 image-text pairs in the training set and 11,738 image-text pairs in the test set.

\begin{table*}[!ht]
    \centering
    \caption{Quantitative Experimental Results of Advanced Methods on the DIOR-IDD Detailed Remote Sensing Image Description Dataset}
    \begin{tabular}{cccccccc}
        \hline
         & \textbf{Methods} & \textbf{LLM Decoder} & \textbf{BLEU-4} & \textbf{METEOR} & \textbf{ROUGE\_L} & \textbf{CIDEr} & \textbf{SPICE} \\ \cline{2-8}
        \multirow{5}{*}{\textbf{Zero-shot}} & InstructBLIP (2023) & Flan-t5-xl \cite{CHL2022} & 2.7 & 11.2 & 16.7 & 2.0 & 10.8 \\
        & InstructBLIP (2023) & Vicuna-7b \cite{vicuna2023} & 2.1 & 10.4 & 17.0 & 1.3 & 8.4 \\
        & LLaVA (2023) & Vicuna-7b & 2.6 & 13.8 & 20.3 & 3.7 & 11.6 \\
        & Vip-LLaVA (2023) & Vicuna-7b & 2.5 & 13.7 & 19.8 & 3.1 & 11.6 \\
        & Kosmos-2 (2023) & Magneto-1.6b \cite{WMH2022} & 3.1 & 13.6 & 20.1 & 3.8 & 12.1 \\ \cline{2-8}
        \multirow{2}{*}{\textbf{Fine-tune}} & LLaVA (2023) & Vicuna-7b & 8.7 & 18.1 & 27.3 & 19.5 & 19.5 \\
        & Vip-LLaVA (2023) & Vicuna-7b & 8.8 & 18.2 & 27.2 & 20.5 & 19.7 \\ \cline{2-8}
        \rowcolor{gray!20} & MGIMM & Phi-2-2.3b & \textbf{13.0} & \uline{18.8} & \textbf{32.6} & \uline{31.5} & \textbf{23.7} \\
        \rowcolor{gray!20} & MGIMM & Vicuna-7b & \uline{9.4} & 18.4 & 27.8 & 24.2 & 20.3 \\
        \rowcolor{gray!20} & MGIMM & Llama-2-7b & \textbf{13.0} & \textbf{19.9} & \uline{31.8} & \textbf{34.6} & \uline{23.3} \\ \hline
        \multicolumn{8}{l}{The first and second best results are marked with \textbf{bold} and \uline{underline}, respectively.} \\
    \end{tabular}
    \label{tab3}
\end{table*}


\subsection{Baseline Methods and Evaluation Metrics}
To evaluate the capability of the method proposed in this paper for providing detailed descriptions of remote sensing images and the effectiveness of region-attribute guided learning, we select four advanced comparative methods: InstructBLIP \cite{DLL2023}, LLaVA \cite{LLL2023}, Kosmos-2 \cite{PWD2023}, and Vip-LLaVA \cite{CLM2023}. Both InstructBLIP and LLaVA are limited to image-level instruction adjustments. Kosmos-2 benefits from pre-training on visual grounding tasks, while Vip-LLaVA involves pre-training with bounding boxes superimposed on images. Both pre-training approaches endow the models with a certain degree of regional awareness.

Given that the training data for the aforementioned methods may not necessarily contain specific knowledge of the remote sensing field, we evaluate the zero-shot performance and fine-tuned performance of these methods on the attribute-guided DIOR-IDD dataset. For the LLaVA model, we embed bounding box information into the detailed text descriptions and then fine-tune the model. For the Vip-LaVA model, we follow its guidelines by drawing the geographic target areas on the images with red bounding boxes before fine-tuning the model.

In terms of evaluation metrics, we opted for BLEU-4 \cite{PRW2002},  \cite{BL2005}, ROUGE\_L \cite{L2004}, CIDEr \cite{VZP2015}, and SPICE \cite{AFJ2016}. BLEU assesses the similarity between the generated text and the reference text by counting the number of overlapping n-grams (ranging from single words to multiple words). METEOR leverages an external dictionary to identify synonyms and perform stemming, offering a more comprehensive assessment of the generated text's quality. ROUGE\_L evaluates the similarity between the generated text and the reference text by comparing the length of the longest common subsequence. CIDEr focuses on measuring the diversity and consistency of the generated descriptions. It calculates scores by comparing word frequency statistics between the generated text and multiple reference texts, emphasizing consistency across different references. Differently from traditional n-gram-based metrics, SPICE aims to capture sparse features in the generated text, such as entities and relationships, using semantic parsing trees to measure the similarity between the generated text and the reference text.

\subsection{Parametric Analysis}
During the training process of MGIMM, we use LoRA for parameter-efficient fine-tuning of the large language model, enabling the model to fit the remote sensing image-text dataset. Compared to fine-tuning all the weights in the pre-trained large language model's weight matrix, LoRA chooses to fine-tune two smaller trainable matrices ($B \in R^{d \times r}, A \in R^{r \times k}, r \ll min\left(d,k\right)$) that approximate updates to the original matrix:

\begin{equation}
    W_{0} + \Delta W = W_{0} + BA,
\end{equation}

\noindent where $W_{0}$ is the pre-trained large language model's weight matrix and $\Delta W$ is the parameter weight update for fine-tuning the large language model. Additionally, during the fine-tuning process, LoRA will scale $\Delta W$ using $\frac{\alpha}{r}$. Therefore, during the training process using LoRA, two hyperparameters, $r$ and $\alpha$, are generated. Based on experience, $\alpha$ is typically set to be twice the value of $r$. A smaller $r$ results in a simpler low-rank matrix, leading to fewer parameters learned during adaptation. This can speed up training and potentially reduce computational requirements.

\begin{figure}[!hb]
  \centering
    \includegraphics[width=\linewidth]{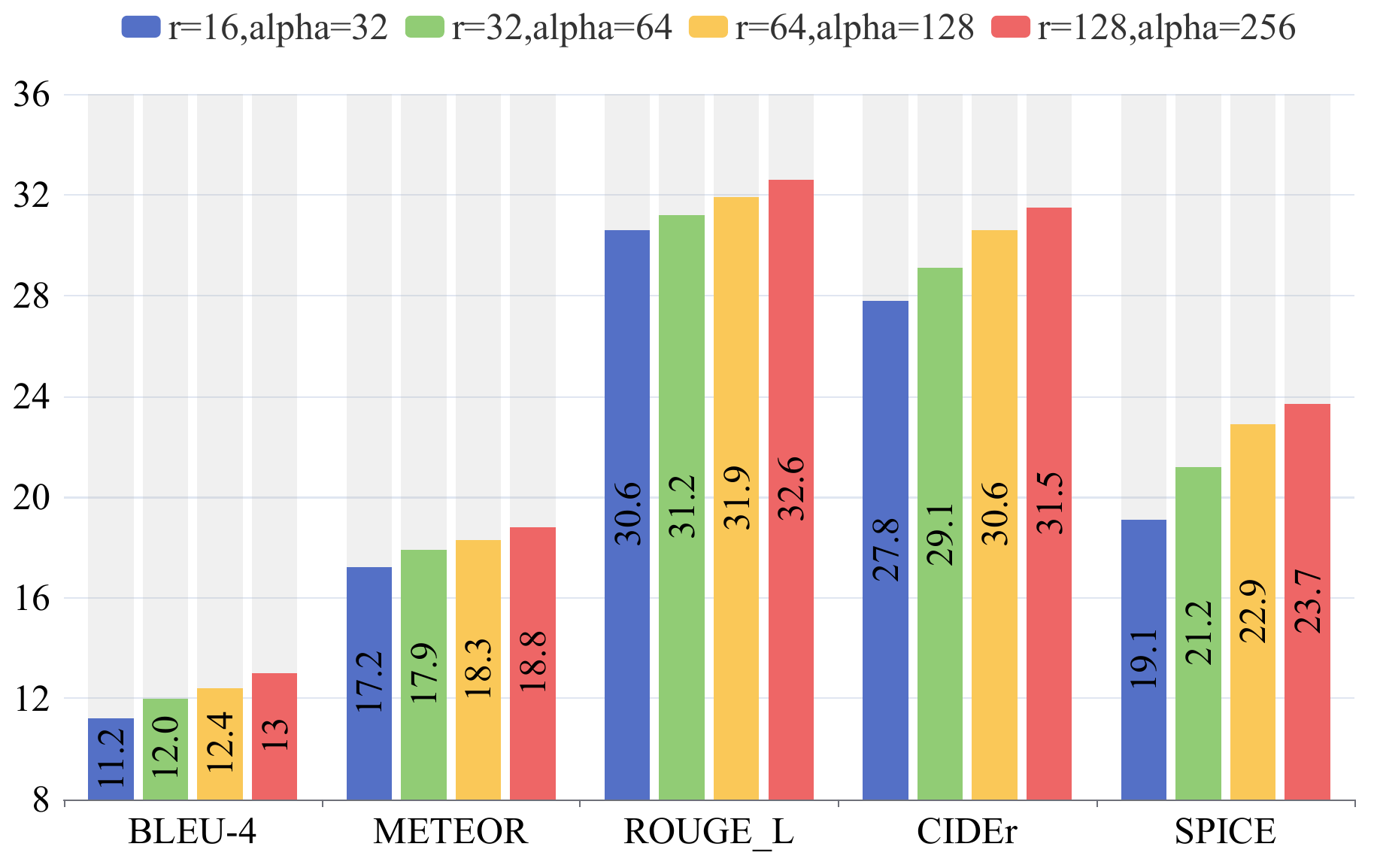}
  \caption{Parameter analysis during the LoRA training process, using the Phi-2 large language model as an example.}
    \label{fig5-a}
\end{figure}

\begin{figure}[!ht]
  \centering
    \includegraphics[width=\linewidth]{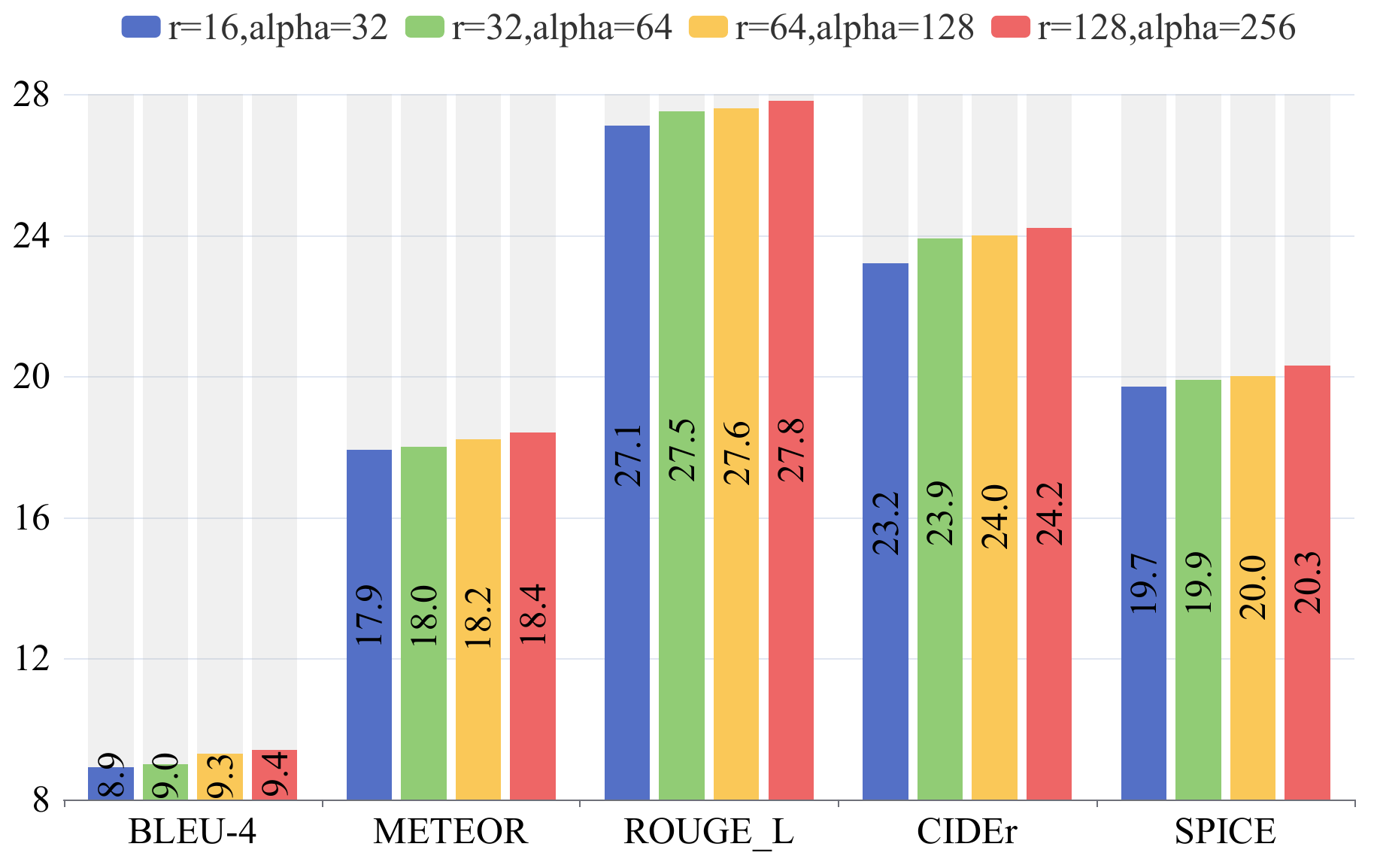}
  \caption{Parameter analysis during the LoRA training process, using the Vicuna large language model as an example.}
    \label{fig5-b}
\end{figure}

\begin{figure}[!ht]
  \centering
    \includegraphics[width=\linewidth]{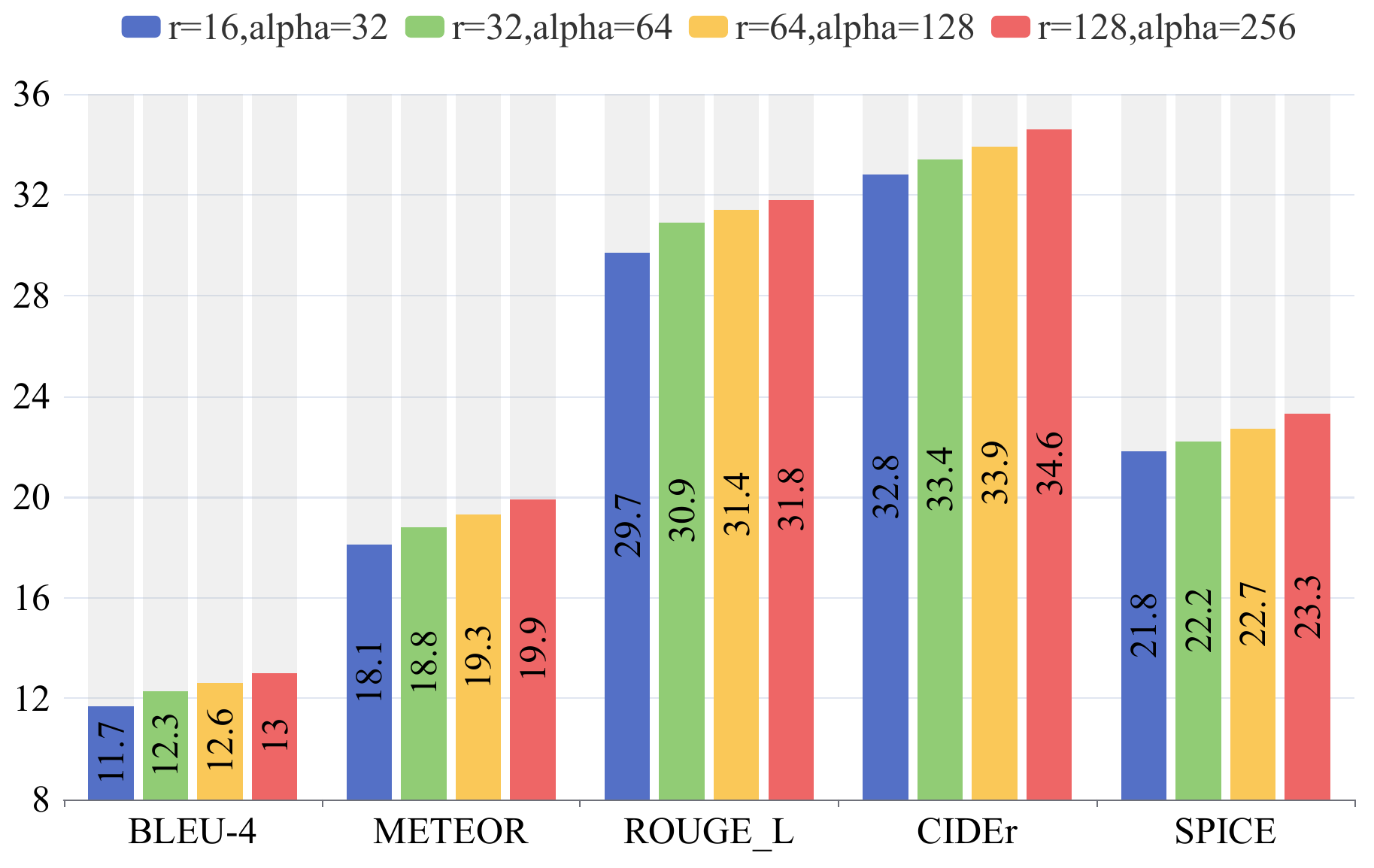}
  \caption{Parameter analysis during the LoRA training process, using the Llama2 large language model as an example.}
    \label{fig5-c}
\end{figure}

We set up four different experiments with varying values of $r$ and $\alpha$, as shown in Fig. \ref{fig5-a}-\ref{fig5-c}. Based on the experimental results, we can see that as $r$ and $\alpha$ decrease, the model's performance gradually declines. This decline is likely due to the reduced ability of the low-rank matrix to capture task-specific information, leading to lower fine-tuning quality. From the experimental result, we set $r$ to 128 and $\alpha$ to 256 in our experiments.

\begin{figure*}[!ht]
  \centering
  \begin{tabular}{c}
       \includegraphics[width=\linewidth]{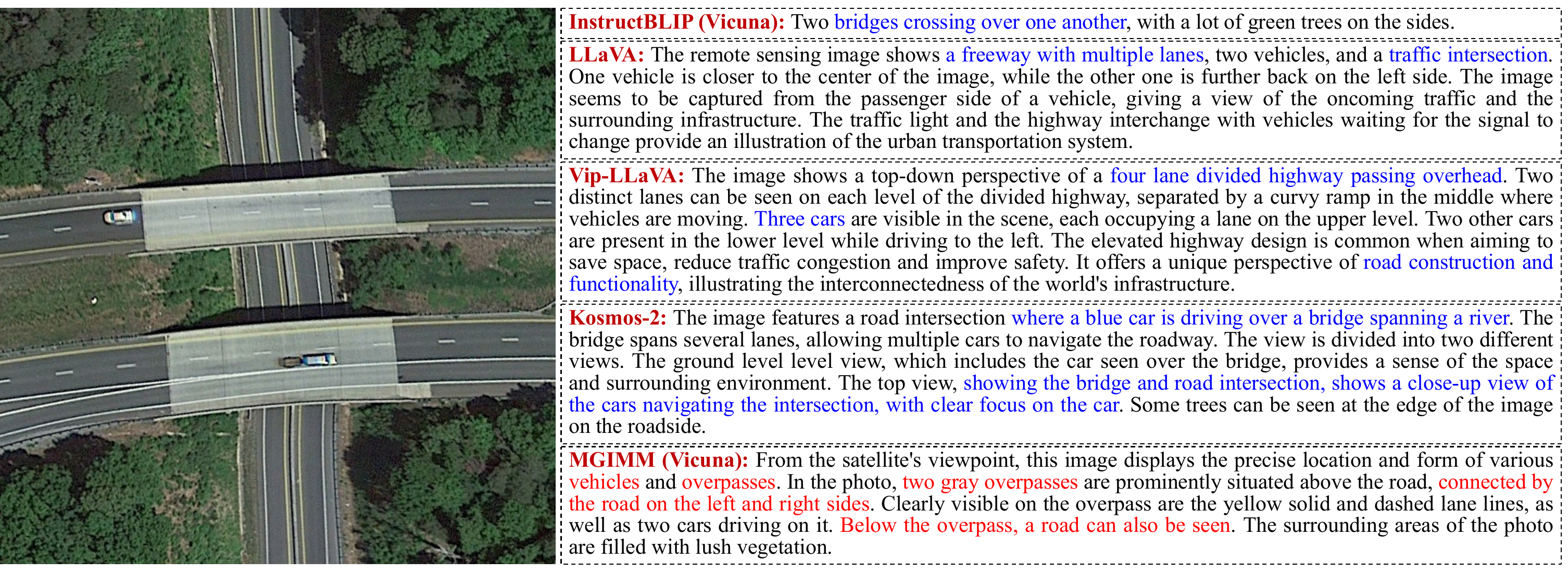}  \\
       (a) \\
       \includegraphics[width=\linewidth]{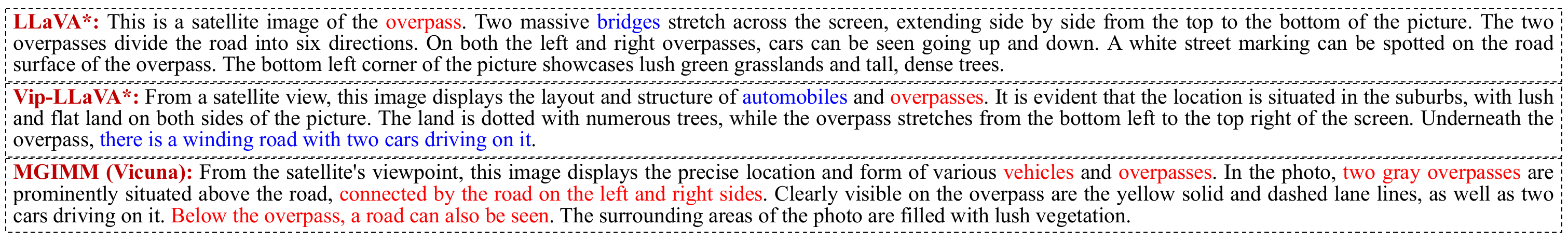}\\
       (b)\\
  \end{tabular}
  \caption{Examples of remote sensing detailed image descriptions on the DIOR-IDD dataset by InstructBLIP (Vicuna), LLaVA, Vip-LLaVA, Kosmos-2, and MGIMM (Vicuna): (a) Results of the zero-shot experiment setup; (b) Results of the fine-tune experiment setup. \textcolor{red}{Red} indicates accurate descriptions, while \textcolor{blue}{blue} signifies erroneous ones.}
    \label{fig6}
\end{figure*}

\subsection{Quantitative Analysis}
Table \ref{tab3} presents the quantitative evaluation results of various methods on the DIOR-IDD dataset, using a range of evaluation metrics. MGIMM surpasses other state-of-the-art (SOTA) comparative methods in performance, a success mainly attributed to region-guided learning. This allows MGIMM to not only align geographic target region-attribute descriptions but also fully leverage the generative capabilities of large language models.

From the zero-shot performance comparison in Table \ref{tab3}, it is clear that methods like InstructBLIP, LLaVA, Vip-LLaVA, and Kosmos-2 lack specialized knowledge in the field of remote sensing. This results in these methods falling behind MGIMM across all evaluation metrics. For instance, the BLEU-4 scores for these methods range from 2.1 to 3.1, significantly lower than the fine-tuned MGIMM, which achieves a BLEU-4 score of 13.0 with Phi-2-2.3b and Llama2-7b decoders. Based on the fine-tuned results, MGIMM remains the best-performing method. Compared to LLaVA, MGIMM significantly outperforms in various evaluation metrics after region-level instruction tuning. The BLEU-4 metric, which measures phrase match accuracy, shows that the phrases generated by MGIMM are closer to the reference text. Additionally, the detailed descriptions generated by MGIMM excel in semantic similarity and structural fluency across other metrics. For example, MGIMM achieves a CIDEr score of 34.6 and a SPICE score of 23.3 with the Llama2-7b decoder, indicating superior descriptive quality and coherence. Compared to Vip-LLaVA, which has regional awareness, MGIMM’s region-level instruction tuning effectively aligns geographic target area-attribute descriptions, improving the quality of long-text descriptions. This is evident from the higher scores in metrics.

Furthermore, we observe that after region-level instruction tuning, Phi-2, with its 2.3 billion parameters, can match the long-text generation capability and quality of Llama2, which has 7 billion parameters, in the context of remote sensing images. This demonstrates that effective region-level tuning can bridge the gap between models of different scales, enabling smaller models to perform on par with larger ones.

Overall, the region-level instruction tuning proposed in this paper significantly reduces the issue of visual ambiguity caused by large-scale variations in remote sensing images. The results underscore the importance of incorporating region-specific knowledge and instruction tuning to enhance the performance of language generative models in the remote sensing domain.

\subsection{Qualitative Analysis}
From the detailed description examples in Figure \ref{fig6}, MGIMM accurately captures the image scenes and concisely describes the position of the overpass in the remote sensing image. Additionally, MGIMM provides detailed descriptions of the lane line colors and the vehicles driving on the road. Compared to the zero-shot models InstructBLIP, LLaVA, and Vip-LLaVA, these methods show a degree of descriptive inaccuracy. For example, InstructBLIP incorrectly describes the ‘overpass’ as a ‘bridge’ and struggles with generating long texts for remote sensing images. LLaVA mistakenly describes the ‘overpass’ as a ‘freeway’ and incorrectly states that there is a ‘traffic intersection’ in the image scene. Vip-LLaVA also wrongly describes the ‘overpass’ as a ‘highway’ and inaccurately reports the number of vehicles in the image. More seriously, Vip-LLaVA’s final sentence contains content that is entirely inconsistent with the remote sensing image scene. Additionally, Kosmos-2 exhibits similar issues to LLaVA and Vip-LLaVA. However, LLaVA, Vip-LLaVA, and Kosmos-2 correctly describe the position of the overpass and the road beneath it.

From the fine-tuning results in Figure \ref{fig6}(b), the descriptive accuracy of LLaVA and Vip-LLaVA shows significant improvement. However, they still have issues with misdescribing geographic targets. For instance, while LLaVA improves in some metrics, it still inaccurately describes the scene elements. More seriously, Vip-LLaVA hallucinates, imagining two cars on the road beneath the overpass, which is not present in the actual image. This hallucination highlights a critical challenge in generating reliable descriptions from remote sensing data. In contrast, MGIMM’s region-level and image-level instruction tuning significantly enhances the model’s ability to align with regional attributes and comprehensively describe key geographic targets in remote sensing images. MGIMM not only avoids the common descriptive errors seen in other models but also ensures that the descriptions are consistent and relevant to the visual content. This results in more accurate and reliable outputs, which are essential for remote sensing applications.

Overall, the results demonstrate that MGIMM, through its progressive tuning strategies, can provide detailed and precise descriptions, effectively overcoming the limitations observed in other models. This reinforces the importance of region-level instruction tuning for improving the performance of language generative models in remote sensing.

\begin{table*}[!ht]
    \centering
    \caption{Ablation Experiment Results of MGIMM}
    \begin{tabular}{ccccccc}
        \hline
        ~ & \textbf{LLM Decoder} & \textbf{BLEU-4} & \textbf{METEOR} & \textbf{ROUGE\_L} & \textbf{CIDEr} & \textbf{SPICE} \\ \cline{2-7}
        MGIMM w/o stage1 \& w/o RIM & Phi-2-2.3b & 8.8 & 18.5 & 27.3 & 21.2 & 19.8 \\
        MGIMM w/o stage1 \& w/ RIM & Phi-2-2.3b & 9.5 & 18.6 & 27.8 & 25.0 & 20.5 \\
        \rowcolor{gray!20} MGIMM & Phi-2-2.3b & 13.0 & 18.8 & 32.6 & 31.5 & 23.7 \\ \cline{2-7}
        MGIMM w/o stage1 \& w/o RIM & Vicuna-7b & 9.1 & 18.1 & 27.5 & 22.1 & 19.8 \\ 
        MGIMM w/o stage1 \& w/ RIM & Vicuna-7b & 8.5 & 17.2 & 26.8 & 19.5 & 18.5 \\
        \rowcolor{gray!20} MGIMM & Vicuna-7b & 9.4 & 18.4 & 27.8 & 24.2 & 20.3 \\ \cline{2-7}
        MGIMM w/o stage1 \& w/o RIM & Llama2-7b & 12.6 & 19.6 & 31.3 & 33.5 & 22.5 \\
        MGIMM w/o stage1 \& w/ RIM & Llama2-7b & 9.6 & 18.5 & 27.9 & 24.4 & 20.4 \\
        \rowcolor{gray!20} MGIMM & Llama2-7b & 13.0 & 19.9 & 31.8 & 34.6 & 23.3 \\ \hline
    \end{tabular}
    \label{tab4}
\end{table*}

\begin{figure}[!ht]
    \centering
    \includegraphics[width=\linewidth]{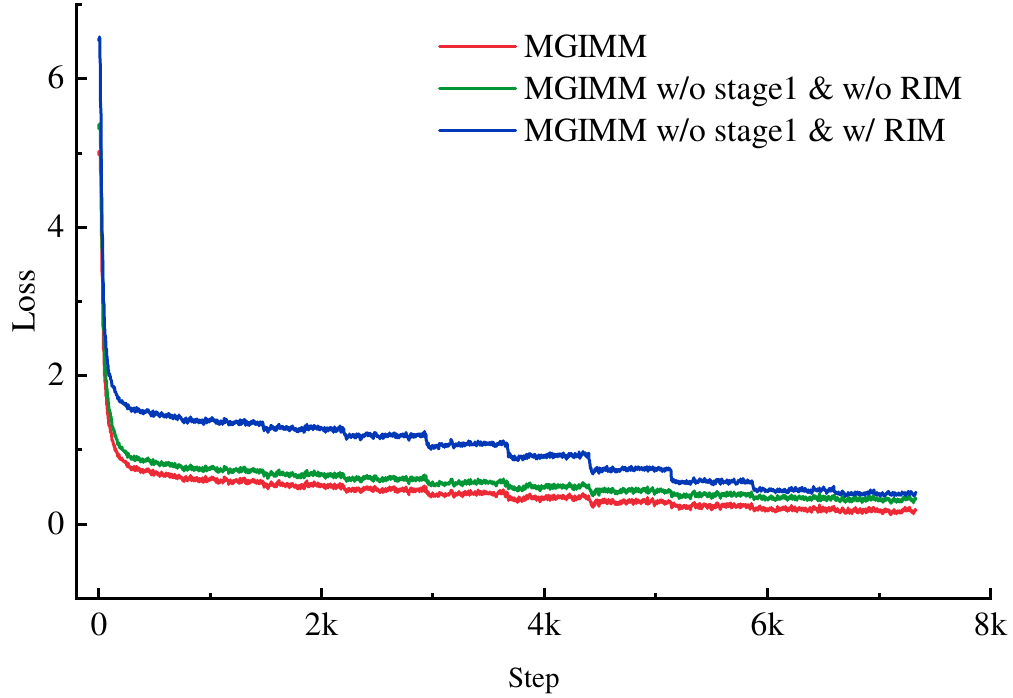}
    \caption{Convergence analysis of two ablation experiments using Llama2 as the large language model.}
    \label{fig7}
\end{figure}

\subsection{Ablation Study}
To further demonstrate the effectiveness of region-level instruction tuning and image-level instruction tuning, we conducted ablation experiments. We set up two ablation experiment configurations: "\textit{MGIMM w/o stage1 \& w/o RIM}," and "\textit{MGIMM w/o stage1 \& w/ RIM}." "\textit{MGIMM w/o stage1 \& w/o RIM}" indicates the use of image-level instruction tuning only. To investigate whether MGIMM can perform end-to-end fine-tuning, we set up an ablation experiment "\textit{MGIMM w/o stage1 \& w/ RIM}." In this experiment, the region interactive module is integrated into the image-level instruction tuning stage. The inputs for this stage consist of global image features, regional interactive features, and image-level instruction embeddings.

The results in Table \ref{tab4} strongly demonstrate the effectiveness of region-level instruction tuning and image-level instruction tuning. When the language decoder is Phi-2 with 2.3 billion parameters, omitting region-level instruction tuning significantly reduces the quality of the detailed descriptions generated by MGIMM for remote sensing images. For instance, the BLEU-4 score drops from 13.0 to 8.8, and the ROUGE\_L score decreases from 32.6 to 27.3, highlighting the critical role of region-level tuning in maintaining high descriptive quality. Similarly, on Vicuna-7b and Llama2-7b models with 7 billion parameters, we observe that the absence of region-level instruction tuning severely impacts MGIMM’s ability to generate detailed descriptions. For the Vicuna-7b model, the BLEU-4 score falls from 9.4 to 8.5, and the CIDEr score decreases from 24.2 to 19.5 without region-level tuning. These declines indicate the significant contribution of region-level instructions to the model’s descriptive capabilities. From the "\textit{MGIMM w/o stage1 \& w/ RIM}" experiment results across three different LLM-Decoder configurations, it is evident that forcing MGIMM’s progressive learning strategy into an end-to-end fine-tuning mode leads to a severe decline in performance. For example, using the Llama2-7b model, the BLEU-4 score drops from 13.0 to 9.6, and the SPICE score decreases from 23.3 to 20.4. A possible reason is that concatenating global image features, regional features, and image-level instruction embeddings causes interference between global and regional features. This interference prevents the model from effectively aligning geographic target areas with their attributes, resulting in persistent visual ambiguity. Moreover, the overall pattern observed across different models suggests that region-level instruction tuning significantly enhances the model’s ability to generate accurate and detailed descriptions, which is crucial for applications in remote sensing where precise information is essential. 

In conclusion, the ablation experiments highlight the necessity of region-level instruction tuning and the detrimental effects of its omission. The results suggest that maintaining a multi-stage, progressive learning approach, rather than forcing an end-to-end fine-tuning mode, is vital for achieving optimal performance in detailed description generation.

\subsection{Convergence Analysis}
The convergence behavior of the algorithm is depicted in Fig. \ref{fig7}, showcasing the loss versus the number of steps for different configurations of the MGIMM model using Llama2 as the large language model. The configurations include the standard "\textit{MGIMM}", "\textit{MGIMM w/o stage 1 \& w/o RIM}", and "\textit{MGIMM w/o stage 1 \& w/ RIM}". It is evident that the MGIMM model demonstrates a clear downward trend in loss as the number of steps increases from Fig. \ref{fig7}, indicating successful convergence. The standard "\textit{MGIMM}" configuration achieves the lowest loss, suggesting that the full implementation of the model, including all stages and the RIM, provides the best performance. In contrast, the "\textit{MGIMM w/o stage 1 \& w/ RIM}" exhibits a higher loss throughout the training process, failing to reach the minimal loss achieved by the standard configuration. This indicates that transforming MGIMM into an end-to-end learning approach causes the model to suffer from interference between regional and global features, leading to slower convergence. Interestingly, the "\textit{MGIMM w/o stage 1 \& w/o RIM}" shows an intermediate performance, with a lower loss than the configuration without both components but still higher than the standard "\textit{MGIMM}". This indicates that the lack of region-level instruction tuning hinders the model's attention to regional geographic features, resulting in slower learning.

In summary, the convergence analysis demonstrates the importance of each component in the MGIMM model. The standard "\textit{MGIMM}" configuration, which includes all stages and the RIM, converges more effectively and achieves the lowest loss. This underscores the synergy between the model’s components, facilitating efficient learning and enhanced performance.

\section{Conclusion}
We propose a two-stage instruction tuning method, MGIMM, to eliminate the visual ambiguities caused by the large-scale variations in remote sensing images. This method enforces alignment of geographic target region-attribute descriptions through region-level instruction tuning and then leverages image-level instruction tuning to make large language models aware of the remote sensing domain, fully utilizing their generative capabilities to descript remote sensing images. Furthermore, we present a novel attribute-guided remote sensing image detailed description dataset, attribute-guided DIOR-IDD. The images are large-scale remote sensing images, and the detailed description texts contain a variety of geographic targets, the relationships among these targets, and interactions between the targets and their background. MGIMM significantly outperforms existing multimodal instruction tuning models when compared with other advanced methods. Moreover, the effectiveness of both region-level and image-level instruction tuning is demonstrated through ablation studies. However, MGIMM is not an end-to-end multimodal instruction tuning model, which makes its tuning process somewhat cumbersome.

\normalem
\bibliographystyle{IEEEtran}
\bibliography{Bibtex/ref.bib}

\newpage

 





\end{document}